\documentclass[11pt,a4paper]{article}
\usepackage[hyperref]{emnlp2018}
\usepackage{times}
\usepackage{latexsym}
\usepackage{amsmath}
\usepackage{amssymb,hyperref}
\usepackage{xspace}
\usepackage{graphicx,array,subcaption}
\usepackage{url,todonotes}

\aclfinalcopy 

\newcommand{\once}[1]{\textbf{#1}}

\newcommand{\dpage}{D-PAGE\xspace}
\newcommand{\model}{D-PAGE-K\xspace}

\title{D-PAGE: Diverse Paraphrase Generation}

\author{Qiongkai Xu, Juyan Zhang, Lizhen Qu\thanks{L. Qu is the corresponding author.}, Lexing Xie, Richard Nock\\
  The Australian National University \\
  Data61 CSIRO\\
  {\{Qiongkai.Xu, Juyan.Zhang, Lizhen.Qu, Lexing.Xie, Richard.Nock\}@anu.edu.au} }

\date{}

\begin{document}
\maketitle
\begin{abstract}
In this paper, we investigate the diversity aspect of paraphrase generation. Prior deep learning models employ either decoding methods or add random input noise for varying outputs. We propose a simple method Diverse Paraphrase Generation (D-PAGE), which extends neural machine translation (NMT) models to support the generation of diverse paraphrases with implicit rewriting patterns. Our experimental results on two real-world benchmark datasets demonstrate that our model generates at least one order of magnitude more diverse outputs than the baselines in terms of a new evaluation metric Jeffrey's Divergence. We have also conducted extensive experiments to understand various properties of our model with a focus on diversity.
\end{abstract}

\section{Introduction}

\textit{Diversity} is an essential characteristic of human language, as the meaning of a text or passage can often be restated.
This work promotes the diversity of language by focusing on \textit{paraphrase generation}, which aims to rephrase a text in multiple ways, while preserving its semantic meaning. Therefore, a full-fledged paraphrase generation system features two desired properties, The first is \textit{fidelity}, preserving semantic meanings while paraphrasing. The second is \textit{diversity}, capturing a wide range of linguistic variations.

In the past few years, deep learning models~\cite{COLING16_DeepParaphrase,Arxiv17Paraphrase_RL} achieve superior performance than the shallow models \cite{Zhao2010LeveragingMM,Wubben2010ParaphraseGA} on paraphrase generation, as well as a related task text simplification~\cite{ACL17_neural_text_simplification,EMNLP17Sentence_simplification_RL}. Those deep models employ the encoder-decoder architecture to formulate paraphrasing as monolingual machine translation problem, where the encoder encodes the input texts into hidden representations, and then the decoder takes the representations as input and generates output word sequences in a word-by-word fashion. 

In recent years, there are growing interests in generating lexically and syntactically diverse texts~\cite{narayan2014hybrid,NAACL2016_DiversityObjFunc,ICCV2017_diverse_image_description}. For deep models, the techniques capable of generating diverse paraphrases fall into two categories: i) applying decoding methods such as sampling or top-$K$ beam search; ii) including random noise~\cite{ICCV2017_diverse_image_description,CVPR17_vae} as model input. The former methods can always be applied to a trained deep model. As the model utilizes the same parameters, serving as a deterministic mapping from an input to one optimal output, the other outputs selected by the decoding methods are suboptimal.  Moreover, empirically, the fidelity of outputs decreases as the number of outputs grows, as observed in Section~\ref{subsec:fedility}. Alternatively, although adding random noise to model inputs can sometimes make slight changes to model outputs, we observe in our experiments that in most times the corresponding models generate identical outputs.

In this paper, we propose a simple but effective method, called \textbf{D}iverse \textbf{PA}raphrase \textbf{GE}neration (\textbf{D-PAGE}), extending neural machine translation (NMT) models to support multiple optimal outputs. We assume that different outputs of the same input can be explained by different rewriting patterns. The rewriting patterns are encoded in different subsets of model parameters, called pattern embeddings. 
In particular, we keep the encoder intact, and augment the decoder with pattern embeddings. Given an input text, the model selects appropriate pattern embeddings for different outputs so that the model uses different parameters for different input-output mappings. Different pattern embeddings may also be used to customize models for specific scenarios. For example, a child education system may prefer a model with relatively small vocabulary while a journalist assistant system prefers richer vocabulary. Our models equipped with pattern embeddings provide more flexibility than the models with fixed parameters.

To measure the diversity between models, caused by varying patterns, we introduce an evaluation metric called Jeffrey's Divergence. According to this metric, our results on two real-world datasets show that our model is at least one order of magnitude better than the baselines. Furthermore, our model with at least one pattern embedding is able to achieve competitive fidelity as the baselines. 
Moreover, since it is difficult to explain the underlying complex rewriting patterns in real-world datasets, we construct synthetic datasets with atomic and interpretable patterns. On those datasets, our model is the only one among all models in comparison being able to capture the predefined patterns.

To sum up, our contributions are three-folds:
\begin{itemize}
	\item To the best of our knowledge, D-PAGE is the first attempt to extend NMT models for diverse outputs from model's perspective.
	\item We propose an evaluation metric and synthetic datasets for measuring diversity caused by various rewriting patterns.
    \item Our model achieves substantial improvement in terms of the new evaluation metric, without compromising fidelity.
\end{itemize}

\section{Related Work}
Our work is closely related to the following three fields,
paraphrase generation, creative language generation, and stylistic language generation.

The mainstream approaches on paraphrase generation are based on monolingual machine translation~\cite{Barzilay2001ExtractingPF,Ibrahim2003ExtractingSP,Zhao2010LeveragingMM,ACL12_sentence_simplification_MT,Coling12_Paraphrasing_for_style,xu2016optimizing}, where the original sequences and paraphrased sequences are treated as source and target languages, respectively. Adopting contemporary Neural Machine Translation~\cite{NIPS14_Seq2Seq} technology, paraphrase generation systems manage to improve quality of paraphrases with better grammaticality and meaning preservation~\cite{ACL17_neural_text_simplification,COLING16_DeepParaphrase,Arxiv17Paraphrase_RL,EMNLP17Sentence_simplification_RL}. Although previous paraphrasing systems provide multiple outputs, they focus on lexical diversity~\cite{ACL12_sentence_simplification_MT,dong2017learning} and syntactic diversity~\cite{narayan2014hybrid,narayan2016paraphrase}. The diversity caused by varying rewriting patterns has been largely neglected.

Recently, there has been increasing attention to the diversity of language generation systems, such as conversation systems~\cite{NAACL2016_DiversityObjFunc,li2016persona} and image caption systems~\cite{AAAI16_SentiCap,CVPR17_StyleNet}.
\newcite{NAACL2016_DiversityObjFunc} introduce a novel loss function to increase the usage of more interesting words. \cite{ACL17_Diversity_Summ} proposes to use a novel attention mechanism to reduce repeating phrases, and thus improve diversity. These two approaches treat the diverse text generation problem as improving creativity of the generator, by introducing lexical diversity. 
Another line of works is capable of generating multiple outputs, by introducing random noise as input of the decoder~\cite{ICCV2017_diverse_image_description} or adding noise to the encoder's final hidden representation~\cite{CVPR17_vae}. 

Our work is also relevant to stylistic language generation that generates sentences with particular writing style. \cite{Coling12_Paraphrasing_for_style} paraphrases sentences with the style of a writer, e.g. Shakespeare. SentiCap~\cite{AAAI16_SentiCap} investigates generating image descriptions with positive or negative sentiments.
Later on, StyleNet \cite{CVPR17_StyleNet} proposes to generate romantic and humorous captions to increase the attractiveness of captions. These approaches can be used to generate outputs with diverse styles, while they rely on the definitions of the styles by human. To train the stylish language generation model, one should collect additional corpora with designated styles, which is expensive in many scenarios. In contrast, our model do not rely on such copora. 


\section{Diverse Paraphrase Generation}
To generate paraphrases with patterns, D-PAGE extends the decoder of NMT models by adding multiple rewriting pattern embeddings as inputs. We apply this method to the widely used sequence-to-sequence (Seq2Seq)~\cite{NIPS14_Seq2Seq}
so that the resulted model, coined \model, is able to generate varying paraphrases according to different rewriting patterns.


Formally, a paraphrase generation model takes a word sequence $X=\langle x_1, x_2, \cdots, x_N \rangle$ of length $N$ as input, where $x_i$ denotes a word from a vocabulary $\mathcal{V}$. The model learns a function to map the input to a paraphrase, $Y=\langle y_1, y_2, \cdots, y_M \rangle$, where $y_j$ is a word also from $\mathcal{V}$. 

\begin{figure*}[ht!]
	\centering
	\includegraphics[width=0.85\linewidth]{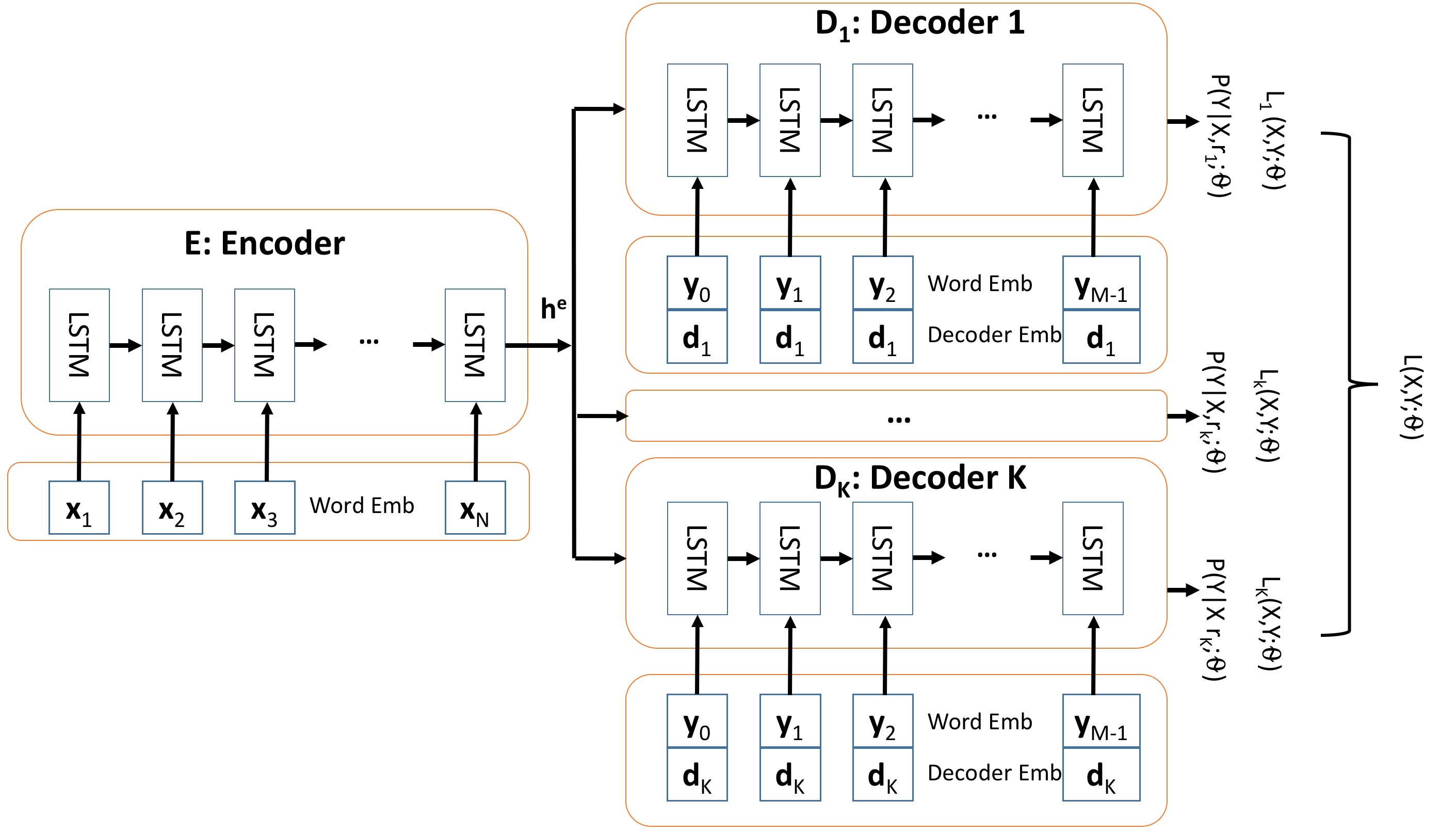}
	\vspace{-1.5mm}
	\caption{The main structure of Diverse Paraphrase Generation method, D-PAGE.}
	\label{fig:system_design}
	\vspace{-5mm}
\end{figure*}

\subsection{Seq2Seq Model}
Seq2Seq architecture consists of two components: (i) an \textit{encoder} projects an input sentence into a representation $\mathbf{h}^e$ and (ii) a \textit{decoder} generates a target word at a time to compose a word sequence. Both of the encoder and the decoder utilize Recurrent Neural Networks (RNN) to construct hidden representations, using Long Short-Term Memory (LSTM)~\cite{Hochreiter1997LongSM}. At the $t$th position, LSTM computes current hidden state $h_{t}$ and memory cell $c_t$ based on an input word embedding $\mathbf{x}_t$, its previous state and memory cell:
\begin{equation}
\mathbf{h}_t, \mathbf{c}_t = \text{LSTM}(\mathbf{h}_{t-1}, \mathbf{c}_{t-1}, \mathbf{x}_t)
\end{equation}
The mapping between the input and the output is defined as a set of the following equations:
\begin{align}
\mathbf{i}_t & =\sigma(\mathbf{W}_i \cdot [\mathbf{x}_t, \mathbf{h}_{t-1}])\\
\mathbf{f}_t & =\sigma(\mathbf{W}_f \cdot [\mathbf{x}_t, \mathbf{h}_{t-1}])\\
\mathbf{o}_t& =\sigma(\mathbf{W}_o \cdot [\mathbf{x}_t, \mathbf{h}_{t-1}])\\
\mathbf{u}_t &=\tanh(\mathbf{W}_u \cdot [\mathbf{x}_t, \mathbf{h}_{t-1}])\\
\mathbf{c}_t &= \mathbf{f}_t \odot \mathbf{c}_{t-1} + \mathbf{i}_t\odot \mathbf{u}_t\\
\mathbf{h}_t &= \mathbf{o}_t \odot \tanh(\mathbf{c}_t)
\end{align}
Conventionally, $\mathbf{i}_t$, $\mathbf{f}_t$, and $\mathbf{o}_t$ are referred to as input gate, forget gate and output gate, respectively.

In this work, the encoder is a (stacked) LSTM which reads a sequence of word embeddings recursively and uses the last hidden state as the final representation of the input $\mathbf{h}^e_N$.

The decoder consists of another (stacked) LSTM, an attention layer and a softmax classifier. The LSTM of the decoder takes the input representation from the encoder as the initial hidden state. At the $t$th step, the hidden state $h_t^d$ is computed by
\begin{equation}
\mathbf{h}_t^d, \mathbf{c}_t^d = \text{LSTM}_{dec}(\mathbf{h}_{t-1}^d, \mathbf{c}_{t-1}^d, \mathbf{y}_{t-1})
\end{equation}
$\mathbf{y}_{t-1}$ denotes the word embeddings of words $y_{t-1}$.
Global attention~\cite{EMNLP15_EffectiveAttention} is added before the softmax layer.
\begin{equation}
a_t(s)=\frac{\exp(\text{score}(\mathbf{h}_t^{d}, \mathbf{h}_s^{e}))}{\sum_{s'}\exp(\text{score}(\mathbf{h}_t^{d}, \mathbf{h}_{s'}^{e}))}
\end{equation}
\begin{equation}
\mathbf{\bar{h}}_t^{d}= \sum_{s} a_t(s) \mathbf{h}_s^{e}
\end{equation}

The softmax classifier computes the probability of each target word $y_t$ as:
\begin{equation}
p(y_t|y_{<t}, x) = \text{softmax}(\mathbf{W}\cdot \mathbf{\bar{h}}_t^d)
\end{equation}


\subsection{\model Model}
There are various ways to rephrase a sentence and preserve its semantic meaning. Different rewriting patterns lead to varying actions. Since those rewriting patterns are not directly observable, \dpage represents them as latent embeddings. Assuming there are $K$ rewriting patterns $\{r_1, r_2, \cdots, r_K\}$, we denote their corresponding pattern embeddings by $\{\mathbf{d}_1, \mathbf{d}_2, \cdots, \mathbf{d}_K\}$.
A decoder is supposed to have different word choices conditioned on different rewriting patterns, thus we feed those pattern embeddings as input of the decoder. We let the pattern embeddings influence the word preference by changing the hidden representations of the decoder through LSTM.
\begin{equation}
\mathbf{h}_t^d, \mathbf{c}_t^d = \text{LSTM}_{dec}(\mathbf{h}_{t-1}^d, \mathbf{c}_{t-1}^d, [\mathbf{y}_{t-1}, \mathbf{d}_k])
\end{equation}
At each time step, $\mathbf{d}_k$ and $\mathbf{y}_{t-1}$ are concatenated to jointly change the hidden states. 
From another perspective, $\mathbf{d}_k$ serves as a pattern condition in addition to previously generated words and input sentences. In this way, a decoder with different rewriting patterns can be viewed as different decoders, as illustrated in Figure~\ref{fig:system_design}.



The LSTM parameters are shared among different decoders in order to capture the common language modeling. Otherwise it can easily lead to the explosion of parameter space. It is also beneficial to learn the pattern embeddings $d_k$ as model parameters instead of using randomly generated ones, as the learned embeddings can better encode the differences between rewriting patterns than the constant noisy ones.

\subsection{Training and Decoding}
The special challenge of training is the unobserved alignment between input sentences and rewriting patterns. We propose a customized loss, which automatically finds the hard assignment of rewriting patterns to input sentences.

For each ground truth paraphrase pair $\langle X, Y \rangle$, the cross entropy loss associated with a rewriting pattern $r_k$ is defined as:
\begin{equation}
\mathcal{L}_k(X, Y;\theta)  = -\sum_{t=1}^{M} \log p(y_t|y_{<t}, X, r_k; \theta)
\end{equation}
The above loss measures the compatibility between input and output sequences using $r_k$. For the same paraphrase pair, if the $\mathcal{L}_k$ is smaller than $\mathcal{L}_i$, $r_k$ is more likely to explain the pair than $r_i$. Thus, we select the smallest loss among all $K$ rewriting patterns for each pair, which is given as:
\begin{equation}
\mathcal{L}(X, Y; \theta) = \min_{k} \mathcal{L}_k(X, Y; \theta)
\end{equation}
Then we apply SGD~\cite{robbins1951stochastic} to update the selected pattern embedding and the model parameters of the backbone Seq2Seq model associated with the smallest loss. 

In the decoding phase, we apply first the encoder to construct the embedding $\mathbf{h}^e$ of an input sequence. Taking $\mathbf{h}^e_N$ as the initial hidden state, each decoder $D_k$ generates the most likely word sequence using beam search, conditioned on its own rewriting pattern. Thus, \model guarantees to generate $K$ sequences.

\section{Experimental Setup}
\subsection{Datasets}
\subsubsection*{Real-world Datasets}
The same as~\cite{COLING16_DeepParaphrase}, we use PPDB~\cite{PPDB_dataset} and Paralex~\cite{Paralex_dataset} for evaluation.

\noindent \textbf{PPDB} is a widely used automatically extracted multilingual paraphrase dataset. We use the most recent \texttt{Phrasal} English PPDB 2.0\footnote{\url{http://paraphrase.org/\#/download}} from the \texttt{XXXL} size. We randomly sample 4,500K and 500K paraphrase pairs for training and testing respectively. 

\noindent \textbf{Paralex} is a large-scale question paraphrase corpus crawled from WikiAnswers\footnote{\url{http://wiki.answers.com}}. Semantically similar questions are annotated by web users. As a result, the dataset contains 5,326,492 question pairs, where 4,826,492 pairs are used for training and the remaining 500K are for testing.~\footnote{We use the same dataset as~\cite{COLING16_DeepParaphrase}}

\subsubsection*{Synthetic Datasets}
Paraphrase generation can be viewed as translating a word sequence to another by applying certain edit operations, such as substitution and insertion. Different choices of edit operation indicate different rewriting patterns.
For real-world sentences, the underlying rewriting patterns are often ambiguous or comprise a complex combination of operations. It is difficult to evaluate which patterns a model fails to capture.
Therefore, we build two synthetic datasets with increasing complexities to evaluate which rewriting patterns a model is able to capture. Each pattern is associated with one or two types of edit operation. The edit operations are aligned with those used in previous works~\cite{narayan2014hybrid,woodsend2011learning}.

\noindent \textbf{Syn-Sub}, one common operation of paraphrasing is to substitute words with their synonyms. As a word could have more than one synonyms, we construct $K$ different dictionaries for the same input words. 
We randomly generate input sequences of length $L \in [6, 20]$.
Then we apply each dictionary to replace all words in an input sequence with their synonyms, and end up with $K$ output sequences. In this way, different dictionaries correspond to different patterns.


\noindent \textbf{Syn-Scale}, another common edit operation is measurement conversion. Herein, each input sequence consists of an integer sampled from $[1000, 10000]$ and a length unit $meter$, such as `2357 $m$'. We convert each input expression to another unit $km$, $dm$, $cm$, $mm$, and $\mu m$ respectively, such as `$2.357\ km$' and `$23570\ dm$'. To avoid a large vocabulary, we treat each digit, each unit of measure, as well as the decimal point as a word. Each conversion corresponds to a rewriting pattern, which comprises two types of operation: i) insertions of digits or the decimal point, and ii) one substitution of unit. 

For both synthetic datasets, we randomly generated 5,000 and 1,000 input sequences for training and testing, separately. Then we transform each of them into five paraphrase sequences, and obtain 25,000 and 5,000 target sequences for training and testing, respectively. 


\subsection{Evaluation}\label{subsec-eval}
We evaluate system performance under two perspectives, fidelity and diversity.

\noindent \textbf{Fidelity} We use multi-reference BLEU~\cite{BLEU}, a modified n-gram precision score, as the main metric. 
We also evaluate the results using SARI~\cite{xu2016optimizing}, which emphasizes the changes in outputs against the inputs\footnote{As SARI provides similar results as BLEU, we provide the results in Appendix~\ref{subsec:sari}.}.

\noindent \textbf{Diversity} 
Diversity caused by different rewriting patterns often appears as different choices of words. \texttt{Distinct-N}~\cite{NAACL2016_DiversityObjFunc} is the only existing evaluation metric for lexical diversity. In particular, let $\mathcal{C}_m$ denote a corpus consisting of all paraphrases generated by a model $m$, \texttt{Distinct-N} is the number of distinct ngrams of order \texttt{N} in $\mathcal{C}_m$ divided by the total number of ngrams of the same order in $\mathcal{C}_m$. This corpus-level measure encourages the generation of a large number of infrequent ngrams. However, a large body of infrequent ngrams often accompany with grammatical errors. Thus, as a common practice in statistics, it is better off measuring differences of word distributions instead of distinct words.

Moreover, it is also desirable to understand the differences between rewriting patterns. Since their differences lead to different word distributions, we apply Kullback-Leibler divergence (KL) \cite{kullback1951information} to characterize the distance between the corresponding word distributions. Let $Q_i$ and $Q_j$ denote the word distributions of a pattern $i$ and a pattern $j$ respectively, we have 
\begin{small}
\begin{equation}
KL(Q_i||Q_j) = \sum_{w\in V} Q_i(w) \log \frac{Q_i(w)}{Q_j(w)} 
\end{equation}
\end{small}
where $V$ denotes the vocabulary. Since there are more than two rewriting patterns, we average KL between all pairs of word distributions as the final measure, which is equivalent to Jeffrey's Divergence (JD)~\cite{jeffreys1998theory}.
\begin{small}
\begin{equation}
JD(Q_1, \cdots, Q_K) = \frac{\sum_{i,j, i\neq j} KL(Q_i||Q_j)}{K(K-1)}
\end{equation}
\end{small}
A high diversity between rewriting patterns corresponds to a large JD, 
which is also a theoretically well grounded measure in information theory and information geometry~\cite{NIPS17_fGANS}. Such a diversity caused by rewriting patterns is referred to as \textit{pattern diversity}.


\subsection{Baselines}
\label{subsec:baselines}
We compare \model with three baselines in terms of both fidelity and diversity:
\begin{itemize}
	\item \textbf{Beam-K} \cite{beam_search} outputs top $K$ paraphrases using beam search.
	\item \textbf{Noise-K} \cite{ICCV2017_diverse_image_description} generates $K$ outputs by adding random noise $z_1, \cdots, z_K$, as input to the decoder.
	\item \textbf{VAE-K} \cite{CVPR17_vae} provides $K$ outputs by introducing noise $z \sim Norm(0, 1)$ as input right after $h_e$, the hidden representations generated by the encoder.
\end{itemize}
Although Noise-K and VAE-K are not explicitly designed for generating sequences with different rewriting patterns, we can still fix the noise for the $k$th decoder and compare the behavior of the model with different noise settings.
Herein, we vary $K$ in $\{2,4,8\}$ for all models in comparison. 

\begin{table*}[t!]
	\centering
	\begin{tabular}{ c|c|c } 
		\hline
		Model & PPDB & Paralex \\ 
		\hline \hline
		Beam-8 & [\textbf{0.165}, 0.148, 0.136, 0.126,& [\textbf{0.350}, 0.317, 0.298, 0.287,\\
		              & 0.120, 0.114, 0.108, 0.103]  & 0.280, 0.276, 0.273, 0.271] \\ \hline
		Noise-8 & [0.163, 0.163, \textbf{0.164}, 0.163,  & [\textbf{0.352}, 0.352,0.352,0.352,\\
		              & 0.163, 0.163, 0.163, 0.163] & 0.352, 0.352, 0.352, 0.352]\\ \hline
		VAE-8    & [\textbf{0.165}, 0.165, 0.165, 0.165, & [\textbf{0.345}, 0.345, 0.345, 0.345,\\
		              & 0.165, 0.165, 0.165, 0.165] & 0.345, 0.345, 0.345, 0.345]\\ \hline 
		\hline
		D-PAGE-2 & [0.138, \textbf{0.167}]                                  & [\textbf{0.347}, 0.329] \\ \hline
        D-PAGE-4 & [0.138, 0.117, 0.145, \textbf{0.154}]                       & [0.343, \textbf{0.350}, 0.302, 0.340] \\ \hline
		D-PAGE-8 & [0.106, 0.134, 0.126, 0.134,& [0.324, 0.341, \textbf{0.342}, 0.332,\\
		                  & \textbf{0.141}, 0.105, 0.137, 0.126] & 0.326, 0.330, 0.296, 0.243]\\
		\hline
	\end{tabular}
    \vspace{-2mm}
    \caption{Multi-reference BLEU of the baselines and D-PAGE-K, on PPDB and Paralex.}
	\label{tab:fidelity_baseline}
\end{table*}

\begin{table}[h!]
	\centering
	\begin{tabular}{ c|c|c } 
		\hline
		Model & PPDB & Paralex \\ 
		\hline \hline
		Seq2Seq\_Layer1 & 0.163 & 0.350 \\ 
		Seq2Seq\_Layer2 & \textbf{0.164} & \textbf{0.351} \\ 
		Seq2Seq\_Layer4 & 0.159 & 0.339 \\
        \hline
    \end{tabular}
    \vspace{-2mm}
	\caption{Comparison of varying Seq2Seq architectures, with multi-reference BLEU.}
	\label{tab:seq2seq_architect}
\end{table}

\section{Experimental Results}
Our extensive experimental results on two real-world datasets show that our model achieves the highest pattern diversity according to our new evaluation metric. At least one decoder gains competitive fidelity as the competitive baselines and half decoders obtain better lexical diversity than the baselines. Our model is also the only one that captures all pre-defined rewriting patterns on the synthetic datasets.

\subsection{Fidelity}
\label{subsec:fedility}
The Seq2Seq model is the backbone of \model and the baselines. We evaluate Seq2Seq with varying number of stacked LSTM layers ($1,2,4$) for both encoder and decoder. As demonstrated in Table~\ref{tab:seq2seq_architect}, Seq2Seq\_Layer2 achieves the highest multi-reference BLEU on both benchmark corpora. We choose Seq2Seq\_Layer2 as the backbone for \model and all three baselines also because 
i) the results are also on a par with those of the same architectures in~\cite{COLING16_DeepParaphrase}; ii) the same architecture is the best performing one in~\cite{ACL17_neural_text_simplification} on text simplification task. 

We provide the BLEU scores of \model and baselines in Table~\ref{tab:fidelity_baseline}, as well as the results of SARI in Appendix~\ref{subsec:sari}.
Compared with the Seq2Seq, at least one decoder of \model achieves competitive BLEU scores, on both PPDB and Paralex. One decoder of D-PAGE-2 even outperforms all Seq2Seq baselines on PPDB. As the number of decoders increases, the averaged BLEU scores of \model decreases. This is due to the fact that the number of training paraphrase pairs assigned to each individual decoder decreases proportionally. A sufficiently large training corpus could thus alleviate such a problem.

The fidelity of the highest ranked sequences generated by Beam-K measure up to those of the backbone model, while the BLEU scores of sequences decrease significantly as their ranks increases.  On average, D-PAGE-8 outperforms Beam-8 Paralex (0.317 vs. 0.294), while keeps competitive on PPDB (0.126 vs. 0.127). Noise-K and VAE-K achieve similar BLEU scores as Seq2Seq, however their generated sequences are mostly identical, because both models are trained to be robust to input noise. 

\begin{table*}[h!]
	\centering
    \small
	\begin{tabular}{ c|ccc|ccc } 
		\hline
		Model & & PPDB & & &Paralex & \\ \cline{2-7}
		& distinct-1 & distinct-2 & JD &distinct-1 & distinct-2  & JD \\
		\hline \hline
		Beam-2       & 0.00380 & 0.05068 & 5.58E-3 & 0.00541 & 0.04464&  1.93E-3\\
		Noise-2      & 0.00331 & 0.03443 & 3.53E-4 & 0.00488 & 0.03350 & 9.06E-5 \\
        VAE-2        & 0.00322 & 0.03298 & 3.27E-5 & 0.00470 & 0.03248 & 3.17E-6\\
		D-PAGE-2     & 0.00353 & 0.04394 & \textbf{1.56E-1} & 0.00490 & 0.03848 & \textbf{2.74E-2} \\
		\hline
		Beam-4        & 0.00214 & 0.03847 & 5.56E-3 & 0.00309 & 0.03119 & 2.29E-3\\
		Noise-4       & 0.00166 & 0.01778 & 3.60E-4 & 0.00247 & 0.01721 & 7.73E-5\\
        VAE-4         & 0.00161 & 0.01654 & 2.48E-5 & 0.00235 & 0.01630 & 2.93E-6\\
		D-PAGE-4      & 0.00188 & 0.02919 & \textbf{5.22E-1} & 0.00263 & 0.02391 & \textbf{4.26E-2} \\
		\hline
		Beam-8       & 0.00109 & 0.02553 & 6.53E-3 & 0.00168 & 0.02123 & 4.07E-3\\
		Noise-8      & 0.00084 & 0.00912 & 3.56E-4 & 0.00124 & 0.00879 & 7.28E-2\\
        VAE-8        & 0.00081 & 0.00830 & 2.31E-5 & 0.00118 & 0.00823 & 5.73E-6\\
		D-PAGE-8     & 0.00101 & 0.02055 & \textbf{5.26E-1} & 0.00142 & 0.01504 & \textbf{5.09E-2} \\
		\hline
	\end{tabular}
    \vspace{-2mm}
    \caption{Distinct-1, distinct-2 and JD of baselines and D-PAGE-K, on PPDB and Paralex.}
	\label{tab:diversity_baseline}
\end{table*}
\hspace{-5mm}

\subsection{Diversity}
\label{subsec:diversity}
We compare the diversity property of \model with baselines in terms of lexical and pattern diversities, using \texttt{distinct-N} and JD respectively.

\texttt{Distinct-N} is a measure of distinct ngrams in generated sequences. From Table~\ref{tab:diversity_baseline}, we can see that both \model and Beam-K outperform Noise-K and VAE-K in terms of this metric. Beam-K achieves the highest \texttt{distinct}, because it deliberately avoids repeating sentences while decoding. However, sequences between different ranks do not tend to capture different rewriting patterns, because their word distributions are similar, as evident by JD scores.

\model obtains dramatically higher JD than Beam-K, Noise-K and VAE-K, which supports the strong information-theoretic argument for diversity in rewriting patterns (Sec~\ref{subsec-eval}). This also demonstrates that the different decoders of \model have their distinct preferences in word usage patterns. With more decoders, JD of \model increases, while there is no such trend for other models.

\begin{table}[h!]
	\centering
	\small
	\begin{tabular}{ c|cc|cc } 
		\hline
		Model &  \multicolumn{2}{c}{PPDB} & \multicolumn{2}{c}{Paralex}\\ \cline{2-5}
         & Beam-2 & Beam-4 & Beam-2 & Beam-4\\
		\hline
		Seq2Seq\_Layer2      & 0.05068 & 0.03847& 0.04464 & 0.03119\\
		\hline
		D-PAGE-2 (D1)        & 0.03501 & 0.02576 & 0.04117 & 0.02800\\
		D-PAGE-2 (D2)        & \textbf{0.06127} & \textbf{0.04581} & \textbf{0.04809} & \textbf{0.03282}\\
		\hline
		D-PAGE-4 (D1)       & 0.04983 & \textbf{0.03889} & \textbf{0.04596} & \textbf{0.03135}\\
		D-PAGE-4 (D2)       & 0.02570 & 0.01919 & 0.03968 & 0.02720\\
		D-PAGE-4 (D3)       & 0.03916 & 0.02907 & \textbf{0.05243} & \textbf{0.03602}\\
		D-PAGE-4 (D4)       & \textbf{0.06237} & \textbf{0.04795} & 0.03621 &0.02490\\
		\hline
	\end{tabular}
	\caption{Comparison of lexical diversity of top-2 and top-4 beam search on Seq2Seq\_Layer2 and \model, using distinct-2.}
	\label{tab:diversity_mixture}
\end{table}

We also apply top-2 and top-4 beam search for D-PAGE-2 (with decoder D1-D2) and D-PAGE-4 (with decoder D1-D4), on both PPDB and Paralex. \texttt{Distinct-2} are reported in Table~\ref{tab:diversity_mixture}. Half of the \model outperform the Seq2Seq\_Layer2 models. Although D-PAGE is not essentially designed for lexical diversity, we can still obtain single models with noticeable higher vocabulary coverage, on both PPDB (D-PAGE-2(D2) and D-PAGE-4(D4)) and Paralex(D-PAGE-4(D3)). 

\begin{table*}[h]
	\centering
	\begin{tabular}{ c|c|c|c|c} 
		\hline
		Dataset  & Model& AvgLen(S) & AvgLen [$D_1$, $D_2$, $D_3$,$D_4$] & $\Delta$ \\\hline \hline
		& Beam-4 &2.438 & [2.584, 2.578, 2.581, \textbf{2.594}] & 0.016\\
		PPDB      & Noise-4 & 2.438 & [\textbf{2.609}, 2.608, 2.606, 2.606] & 0.003\\
        & VAE-4 & 2.438 & [2.616, \textbf{2.617}, 2.616, 2.616] & 0.001\\
		& D-PAGE-4 & 2.438 & [2.516, \textbf{3.315}, 2.765, 2.084] & \textbf{1.231}\\
		\hline \hline
		& Beam-4 & 6.643 & [5.844, 5.870, 5.904, \textbf{5.965}] & 0.121\\
		Paralex   & Noise-4 & 6.643 & [5.760, \textbf{5.763}, 5.756, 5.758] & 0.007\\
        & VAE-4 & 6.643 & [5.852, 5.852, \textbf{5.853}, 5.852] & 0.001\\
		& D-PAGE-4 & 6.643 & [5.961, 6.459, 5.270, \textbf{6.532}] & \textbf{1.262}\\
		\hline
	\end{tabular}
	\caption{The average length of the sources, AvgLen(S), and average length of outputs of $k$th decoder, $\text{AvgLen}(D_k)$, across Beam-4, Noise-4, VAE-4 and D-PAGE-4, on PPDB and Paralex.}
	\label{tab:diversity_length}
\end{table*}

\begin{table*}[h]
	\centering
    \small
	\begin{tabular}{ |l|l|l|l|}
    	\hline
		sr:\ responses to  &  sr:\ am pleased & sr:\ managed to & sr:\ partly or entirely \\ \hline
        $D_1$: responses provided & $D_1$: am happy & $D_1$: was able to & $D_1$: partially or totally \\
        $D_2$: replies to the & $D_2$: am very happy to & $D_2$: have been able to & $D_2$: partially or wholly \\
        $D_3$: the replies to & $D_3$: is my pleasure to & $D_3$: been able to
& $D_3$: either partially or totally \\
        $D_4$: answers to & $D_4$: am happy & $D_4$: managed to
 & $D_4$: partially or totally \\
		\hline
	\end{tabular}
\hspace{-2mm}
	\caption{Sample results, with sources (sr) and \dpage outputs of the $k$th decoder $D_k$, on PPDB.}
	\label{tab:case_study}  
\end{table*}
\hspace{-10mm}

\subsection{Analysis}
\label{subsec:analysis}
We investigate \model in details by using its generated sequences based on PPDB and Paralex. Inspired by the work of authorship analysis~\cite{Diederich2003AuthorshipAW} and language style classification~\cite{khosmood2008automatic}, we analyse the average output length and the preferences of function words. We also observe the performance of \model on synthetic datasets and demonstrate two types of rewriting patterns that can be learned by our model. 

Table~\ref{tab:diversity_length} lists the average length of input sentences ($\text{AvgLen}(S)$) and average length of the outputs of $k$th decoder ($\text{AvgLen}(D_k)$) for all models. We also calculate the differences between longest and shortest averaged outputs by $\Delta=\max_k \text{AvgLen}(D_k)-\min_k \text{AvgLen}(D_k)$.
The average length of output sequences using \model is indeed longer than those of baselines. \model also achieves the largest $\Delta$ on both PPDB and Paralex, larger than 1.2. This shows that \model encodes more diverse patterns, with some decoders prefer to generate longer, probably more complex, sentence than others.
In contrast, the baselines do not feature such distinct properties.

Table~\ref{tab:case_study} lists sample outputs of \model with $K = 4$ on PPDB, together with source sentences (sr). In most cases, \model is able to replace the words with synonyms, e.g. ``responses'' to ``replies'' and ``answers'', ``entirely'' to ``wholly'' and ``totally''. The tense of the verbs in output phrases are also coherent with the inputs, e.g. ``am'' to ``is'', ``managed'' to ``was''. In some cases, our model generates duplicated outputs. For example, the output of ``managed to'' through $D_4$ is the same as the input. This is due to the limitation of Seq2Seq models, which does not guarantee the dissimilarity between the outputs and inputs. The outputs of ``partly or entirely'' through $D_1$ and $D_4$ are duplicated, which means that even through the models with different rewriting patterns, the outputs can still be the same.

For each decoder, we demonstrate the top 10 words contributed to JD, in Table~\ref{tab:top_word}. For both PPDB and Paralex, function words contribute most, such as articles, auxiliary verbs, prepositions, conjunctions, etc. Different decoders of \model trained on Paralex show preferences of different question words, as Parelex is composed of question paraphrases. As the selection bias of function words was used in text style analysis~\cite{Diederich2003AuthorshipAW}, we believe such word preference between decoders illustrates the patterns of decoders, to some extent.

\begin{table}[t]
	\centering
    \small
    \begin{tabular}{l}
        \hline
        PPDB\\ \hline
        D$_1$: \once{was}, are, be, \once{its}, \once{these}, \once{those}, \once{under}, \once{during}, \once{within}, \once{all} \\
        D$_2$: the, \once{out}, been, be, \once{a}, \once{set}, has, \once{up}, to, are\\
        D$_3$: is, been, the, to, \once{being}, of, \once{shall}, \once{had}, \once{have}, has\\
        D$_4$: is, \once{were}, \once{by}, \once{and}, \once{in}, of, \once{their}, \once{on}, \once{such}, \once{that}\\
        \hline
        \hline
        Paralex\\ \hline
        D$_1$: do, what, in, \once{where}, be, \once{you}, s, how, \once{and}, for\\
        D$_2$: the, a, of, some, between, any, \once{history}, main, most, \once{all}\\
        D$_3$: be, how, what, do, \once{can}, s, for, in, \once{who}, \once{to}\\
        D$_4$: the, a, of, some, between, \once{different}, any, main, \once{name}, most\\
		\hline
	\end{tabular}
	\vspace{-2mm}
    \caption{Top 10 words contribute most to JD, for each decoder $D_k$, on PPDB and Paralex. Words that appear once in the lists are \once{bold}.}
	\label{tab:top_word}
    \vspace{-4mm}
\end{table}

In order to understand the capability of capturing specific rewriting patterns, we conduct experiments on synthetic datasets for our model. In particular, we apply $K$ decoders to generate $K$ sets of paraphrases. Since we have a reference set per rewriting pattern, we pair each paraphrase set with each reference set. For each pair, we align each system output to a single reference, so that we are able to compute a BLEU score for that pair. The corresponding results are reported in Table~\ref{tab:syn_D-PAGE}. If a model is able to capture all patterns, we expect that for each pattern, there is one decoder achieving a high score.
For Syn-Sub, our model is able to learn the patterns $r_1$, $r_3$ and $r_4$, but only partially capture the pattern $r_2$. Our further investigation shows that $D_5$ tends to provide outputs with mixed rewriting pattern between $r_2$ and $r_4$ . 
For Syn-Scale, although the lengths of input and output sequences are different and the dataset include two types of operations, substitution and insertion, our model can still learn all possible patterns perfectly. We contribute this to smaller vocabulary size and shorter sequence length.
However, we have not observed such distinct rewriting pattern from the outputs of Noise-K or VAE-K, see Appendix~\ref{sec:rand_syn}.

\begin{table}[!t]
	\centering
    \small
	\begin{subtable}{1.0\linewidth}
		\centering
		\begin{tabular}{cccccc}
			\hline
			Dec & \multicolumn{5}{c}{References} \\
			\cline{2-6}
			Id& r1 & r2 & r3 & r4 & r5 \\
			\hline
			$D_1$ &0.000&0.000&0.000&\textbf{1.000}&0.000\\ \hline
			$D_2$ &0.000&0.000&\textbf{1.000}&0.000&0.000\\ \hline
			$D_3$ &\textbf{1.000}&0.000&0.000&0.000&0.000\\ \hline
			$D_4$ &0.000&0.000&\textbf{1.000}&0.000&0.000\\ \hline
			$D_5$ &0.000&0.369&0.000&\textbf{0.631}&0.000\\ \hline
		\end{tabular}
		\caption{Syn-Sub}
	\end{subtable}
	
	
	\begin{subtable}{1.0\linewidth}
		\centering
		\begin{tabular}{cccccc}
			\hline
			Dec & \multicolumn{5}{c}{References}\\
			\cline{2-6}
			Id& r1 & r2 & r3 & r4 & r5\\
			\hline
			$D_1$ &0.047&0.577&0.719&\textbf{1.000}&0.612\\ \hline
			$D_2$ &0.066&\textbf{1.000}&0.693&0.598&0.386\\ \hline
			$D_3$ &0.055&0.671&\textbf{1.000}&0.736&0.502\\ \hline
			$D_4$ &0.033&0.407&0.507&0.607&\textbf{1.000}\\ \hline
			$D_5$ &\textbf{1.000}&0.066&0.057&0.049&0.032\\ \hline
		\end{tabular}
		\caption{Syn-Scale}
	\end{subtable}
    \vspace{-2mm}
	\caption{The confusion matrices of D-PAGE on Syn-Sub and Syn-Shift using BLEU scores.}
	\label{tab:syn_D-PAGE}
\vspace{-4mm}
\end{table}

\section{Conclusion}
This paper presents a novel paraphrase generation method \dpage, which generates paraphrases with multiple rewriting patterns. Our model with different patterns can generate diverse outputs, with little loss of fidelity. \dpage also manages to provide models with higher lexical diversity than baselines. The experiments on synthetic datasets demonstrate the strong capacity of our model to learn common rewriting patterns.

\bibliography{emnlp2018}
\bibliographystyle{acl_natbib_nourl}

\newpage
\ 
\newpage
\appendix


\section{Training Balance of D-PAGE}
\label{sec:train_balance}
\begin{figure}[h]
	\begin{subfigure}{1.1\linewidth}
		\centering
		\includegraphics[width=1.\linewidth]{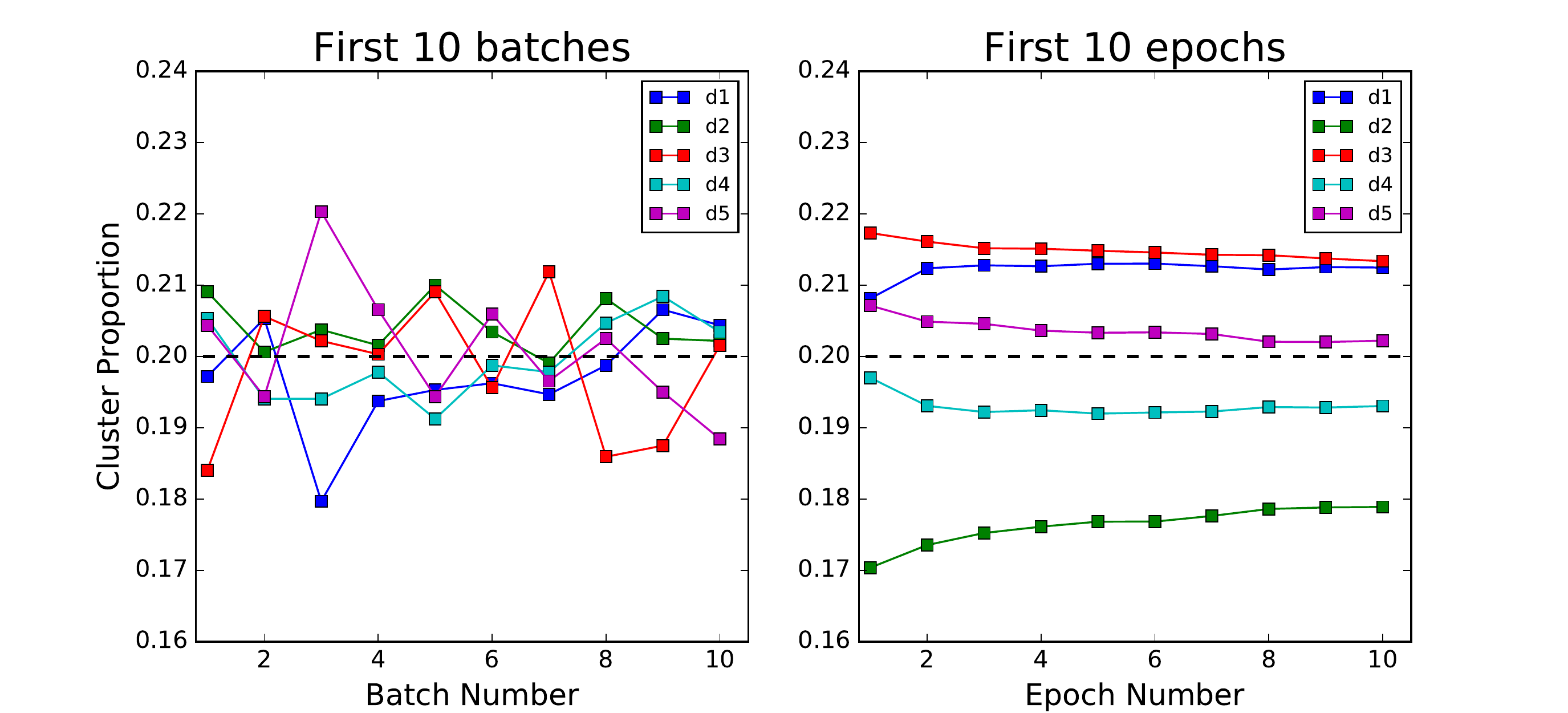}
		\caption{PPDB}
		\label{fig:propotion_ppdb}
	\end{subfigure}
    
	\begin{subfigure}{1.1\linewidth}
		\centering
		\includegraphics[width=1.\linewidth]{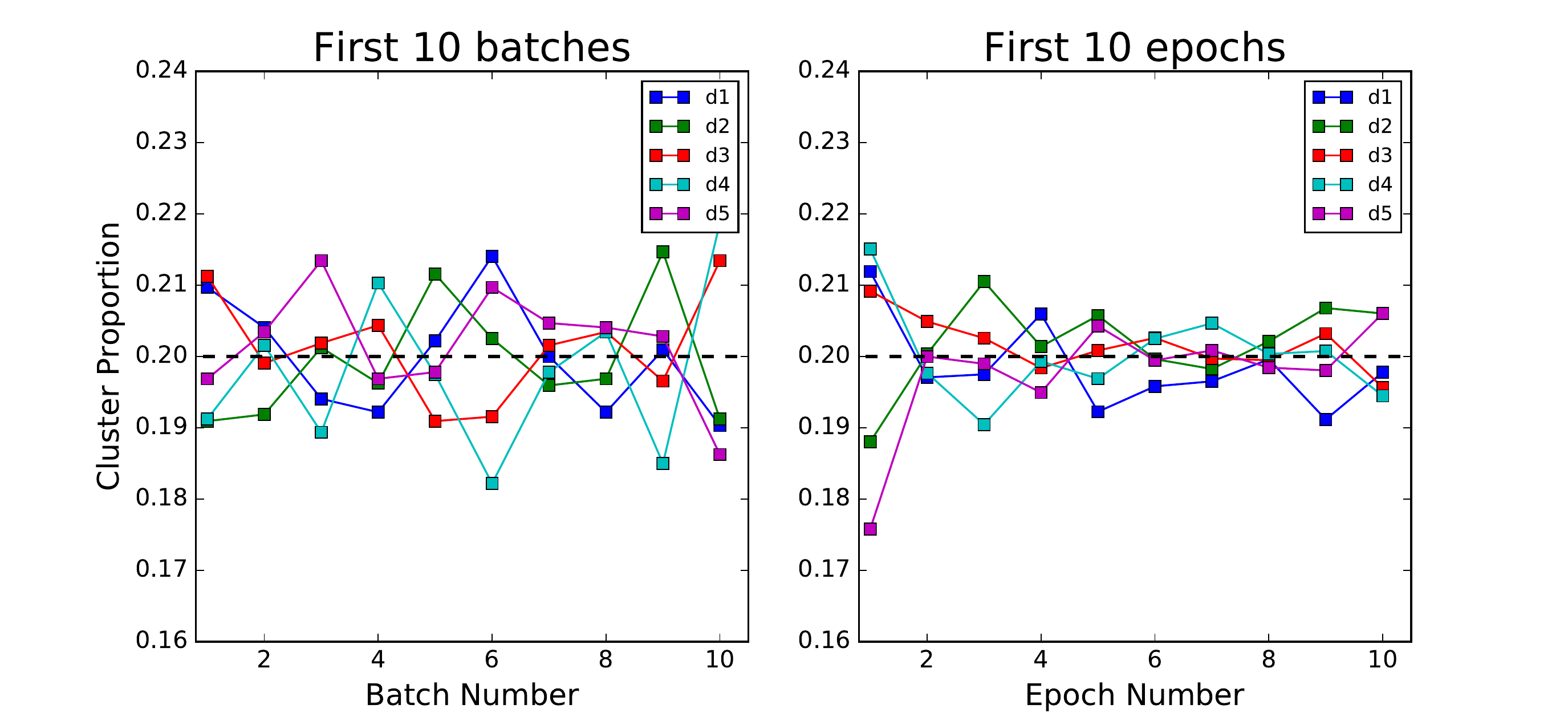}
		\caption{Paralex}
		\label{fig:propotion_paralex}
	\end{subfigure}
    \caption{Proportions of training samples get the lowest loss through the $k$th decoder, for the first 10 batches and epochs, on PPDB and Paralex.}
    \label{fig:propotion}
\end{figure}

We demonstrate the proportions of training samples that get the lowest loss (match best) through  the $k$th decoder, $d_k$, see Figure~\ref{fig:propotion}. During the first 10 batches, the proportions vibrate sharply. After several epochs, the proportions gradually converge to the value close to 0.2 (the ideal averaged proportion). This result shows that the training samples fall into each decoder with balance. As PPDB is composed of shorter sequences than Paralex, the clustering proportions of PPDB converge better.

\section{Efficiency of D-PAGE}
\label{sec:efficiency}
We develop our \model model based on OpenNMT \cite{opennmt}. We conduct all our experiments on the server with one Nvidia Tesla P100 (SXM2). For all experiments, we train the models for 10 epochs. The total training time of Seq2Seq, with 2 layers, and D-PAGE, with 2-8 decoders, are provided in hour(h), in Table~\ref{tab:efficiency}.

Introducing more decoders linearly increases the time for training. Paralex costs roughly 2 times more training time than PPDB, as the average sentence length of Paralex is longer than that of PPDB. The training time of D-PAGE-8 for PPDB and Paralex are acceptable, which are less than 1.5 days and  4.1 days respectively. As D-PAGE models share the calculation of encoder part, training time of D-PAGE-8 is about 5.5 times more than single decoder models and 2 times more than D-PAGE-2. The training efficiency can be improved by calculating the loss of each decoder in parallel through multiple GPUs.
For decoding, each decoder for D-PAGE works as efficiently as Seq2Seq's decoder.

\begin{table}[t!]
	\centering
    \small
	\begin{tabular}{ c|l|l|l|l}
    	\hline
        Dataset & Seq2Seq & D-PAGE-2 & D-PAGE-4 & D-PAGE-8\\
        \hline
        PPDB    & \ \ 5.20h & 10.00h & 18.27h & 33.12h\\
        \hline
        Paralex & 14.99h  & 29.97h & 53.40h & 97.90h \\
		\hline
	\end{tabular}
	\caption{Training time consumptions of D-PAGE-K and Seq2Seq, on PPDB and Paralex.}
	\label{tab:efficiency}
\end{table}

%

\begin{table*}[ht!]
	\centering
	\begin{tabular}{ c|c|c } 
		\hline
		Model & PPDB & Paralex \\ 
		\hline \hline
		Beam-8 & [\textbf{0.225}, 0.222, 0.221, 0.221,& [\textbf{0.447}, 0.441, 0.438, 0.436,\\
		              & 0.220, 0.221, 0.220, 0.220] & 0.435, 0.434, 0.435, 0.435] \\ \hline
		Noise-8 & [\textbf{0.225}, 0.225, 0.225, 0.225,  & [\textbf{0.450}, 0.450, 0.450, 0.450,\\
		              & 0.225, 0.225, 0.225, 0.225] & 0.450, 0.450, 0.450, 0.450]\\ \hline
		VAE-8    & [\textbf{0.224}, 0.224, 0.224, 0.224, & [\textbf{0.446}, 0.446, 0.446, 0.446,\\
		              & 0.224, 0.224, 0.224, 0.224] & 0.446, 0.446, 0.446, 0.446]\\ \hline 
		\hline
        D-PAGE-4 & [0.213, 0.220, \textbf{0.228}, 0.221] & [0.446, \textbf{0.449}, 0.435, 0.443] \\ \hline
		D-PAGE-8 & [0.215, 0.225, 0.215, 0.217, & [0.443, 0.449, \textbf{0.449}, 0.447,\\
		                  & 0.223, \textbf{0.226}, 0.208, 0.209] & 0.439, 0.444, 0.431, 0.418]\\
		\hline
	\end{tabular}
	\caption{Fidelity experimental results on PPDB and Paralex, with SARI.}
	\label{tab:fidelity_baseline_sari}
\end{table*}


\section{SARI on PPDB and Paralex}
\label{subsec:sari}
The experiment result of SARI on PPDB and Paralex is demonstrated in Table~\ref{tab:fidelity_baseline_sari}. The overall results are similar to those of multi-reference BLEU, as discussed in Section~\ref{subsec:fedility}. Surprisingly, in terms of SARI, both of the best D-PAGE-2 and D-PAGE-4 models outperform the best Beam-K demonstrates the potential of using D-PAGE to train model with higher fidelity.

\section{Baseline results on Synthetic Datasets}
\label{sec:rand_syn}
In this section, we demonstrate the performance of Noise-K and VAE-K on the Synthetic datasets, in Table~\ref{tab:syn_noise} and Table~\ref{tab:syn_vae}.
There are mainly two types of behavior for both Noise-K and VAE-K, on our synthetic datasets. i) The models totally get lost among different rewriting patterns, and fail to generate coherent results, such as Noise-K on Syn-Sub and Syn-Scale and VAE-K on Syn-Sub. ii) The models manage to learn one of the rewriting patterns, while they ignore other possible ways to paraphrase, such as VAE-K on Syn-Scale.  Comparing with the results of D-PAGE, in Table~\ref{tab:syn_D-PAGE}, D-PAGE is able to detect and learn diverse rewriting operations, such as replacements and insertions.

\begin{table}[!h]
	\centering
	\begin{subtable}{1.0\linewidth}
		\centering
		\begin{tabular}{cccccc}
			\hline
			Dec & \multicolumn{5}{c}{References} \\
			\cline{2-6}
			Id& r1 & r2 & r3 & r4 & r5 \\
			\hline
			D1 &0.030&0.012&0.009&\textbf{0.047}&0.006\\ \hline
			D2 &0.031&0.011&0.009&\textbf{0.047}&0.004\\ \hline
			D3 &0.031&0.011&0.009&\textbf{0.048}&0.006\\ \hline
			D4 &0.030&0.011&0.008&\textbf{0.046}&0.006\\ \hline
			D5 &0.030&0.011&0.010&\textbf{0.046}&0.004\\ \hline
		\end{tabular}
		\caption{Syn-Sub}
	\end{subtable}
	

	\begin{subtable}{1.0\linewidth}
		\centering
		\begin{tabular}{cccccc}
			\hline
			Dec & \multicolumn{5}{c}{References} \\
			\cline{2-6}
			Id& r1 & r2 & r3 & r4 & r5 \\
			\hline
			D1 &\textbf{0.025}&0.007&0.006&0.005&0.003\\ \hline
			D2 &\textbf{0.022}&0.004&0.004&0.003&0.002\\ \hline
			D3 &\textbf{0.027}&0.006&0.006&0.005&0.003\\ \hline
			D4 &\textbf{0.025}&0.006&0.005&0.005&0.003\\ \hline
			D5 &\textbf{0.018}&0.004&0.003&0.003&0.002\\ \hline
		\end{tabular}
		\caption{Syn-Scale}
	\end{subtable}
	\caption{The confusion matrices of Noise-5 on Syn-Sub and Syn-Scale, using BLEU.}
	\label{tab:syn_noise}
\end{table}

\begin{table}[!h]
	\centering
	\begin{subtable}{1.0\linewidth}
		\centering
		\begin{tabular}{cccccc}
			\hline
			Dec & \multicolumn{5}{c}{References} \\
			\cline{2-6}
			Id& r1 & r2 & r3 & r4 & r5 \\
			\hline
			D1 &0.184&\textbf{0.263}&0.204&0.231&0.119\\ \hline
			D2 &\textbf{0.265}&0.192&0.205&0.207&0.131\\ \hline
			D3 &\textbf{0.285}&0.166&0.186&0.211&0.152\\ \hline
			D4 &0.233&0.196&0.174&\textbf{0.234}&0.163\\ \hline
			D5 &0.202&0.193&0.164&\textbf{0.264}&0.178\\ \hline
		\end{tabular}
		\caption{Syn-Sub}
	\end{subtable}
	

	\begin{subtable}{1.0\linewidth}
		\centering
		\begin{tabular}{cccccc}
			\hline
			Dec & \multicolumn{5}{c}{References} \\
			\cline{2-6}
			Id& r1 & r2 & r3 & r4 & r5 \\
			\hline
			D1 &0.899&0.907&\textbf{0.911}&0.904&0.882\\ \hline
			D2 &0.896&0.907&\textbf{0.912}&0.904&0.881\\ \hline
			D3 &0.899&0.907&\textbf{0.910}&0.903&0.882\\ \hline
			D4 &0.899&0.908&\textbf{0.911}&0.903&0.881\\ \hline
			D5 &0.900&0.909&\textbf{0.912}&0.903&0.880\\ \hline
		\end{tabular}
		\caption{Syn-Scale}
	\end{subtable}
	\caption{The confusion matrices of VAE-5 on Syn-Sub and Syn-Scale, using BLEU.}
	\label{tab:syn_vae}
\end{table}

\end{document}